\documentclass[lettersize,journal,twoside]{IEEEtran}
\usepackage{amsmath,amsfonts,amssymb}
\usepackage{algorithm}
\usepackage{algpseudocode}
\usepackage{array}
\usepackage[caption=false,font=normalsize,labelfont=sf,textfont=sf]{subfig}
\usepackage{textcomp}
\usepackage{stfloats}
\usepackage{url}
\usepackage{verbatim}
\usepackage{graphicx}
\usepackage{cite}
\usepackage[export]{adjustbox}
\usepackage[table,xcdraw]{xcolor}
\usepackage{color}
\usepackage[hidelinks]{hyperref}
\definecolor{linkcolor}{HTML}{ee0000}
\hypersetup{
     colorlinks=true,
	 linkcolor=linkcolor,
	 filecolor=linkcolor,
	 urlcolor=linkcolor,
	 citecolor=linkcolor
}
\usepackage{multirow}
\usepackage{booktabs}
\usepackage{orcidlink}
\usepackage{threeparttable}
\usepackage{pifont}

\newcommand\mlcomment[1]{\iffalse #1 \fi}
\newcommand\bsm[1]{\boldsymbol{\mathrm{#1}}}
\newcommand\transform[2]{{\bsm{T}_{#1}^{#2}}}
\newcommand\transformhat[2]{{\hat{\bsm{T}}_{#1}^{#2}}}

\newcommand\rotation[2]{{\bsm{R}_{#1}^{#2}}}
\newcommand\rotationhat[2]{{\hat{\bsm{R}}_{#1}^{#2}}}

\newcommand\translation[2]{{\bsm{p}_{#1}^{#2}}}
\newcommand\translationhat[2]{{\hat{\bsm{p}}_{#1}^{#2}}}

\newcommand\smallminus{{\text{-}}}
\newcommand\smallplus{{\text{+}}}
\newcommand\coordframe[1]{\underrightarrow{\mathcal{F}}_{#1}}
\newcommand{\tabtitlespace}{\vspace{-5pt}}

\newcommand{\figsize}{0.95}
\newcommand{\tabwidth}{10pt}
\newcommand{\equabovespace}{\setlength{\abovedisplayskip}{2pt}}
\newcommand{\equbelowspace}{\setlength{\belowdisplayskip}{2pt}}
\newcommand{\figtabbottomspace}{\vspace{-15pt}}
\usepackage{siunitx}
\begin{document}

\title{eKalibr: Dynamic Intrinsic Calibration for Event Cameras From First Principles of Events}
\author{

Shuolong Chen \hspace{-1mm}$^{\orcidlink{0000-0002-5283-9057}}$, Xingxing Li \hspace{-1mm}$^{\orcidlink{0000-0002-6351-9702}}$, Liu Yuan \hspace{-1mm}$^{\orcidlink{0009-0003-6039-7070}}$, and Ziao Liu \hspace{-1mm}$^{\orcidlink{0009-0009-6644-0272}}$

\thanks{
This work was supported by the the National Science Fund for Distinguished Young Scholars of China under Grant 42425401.}
\thanks{The authors are with the School of Geodesy and Geomatics (SGG), Wuhan University (WHU), Wuhan 430070, China.
Corresponding author: Xingxing Li (\texttt{xxli@sgg.whu.edu.cn}). 
The specific contributions of the authors to this work are listed in Section \hyperref[sect:author_contribution]{\textbf{CRediT Authorship Contribution Statement}} at the end of the article.}

}

\markboth{Journal of \LaTeX\ Class Files,~Vol.~14, No.~8, August~2021}
{Chen \MakeLowercase{\textit{et al.}}: eKalibr: Dynamic Intrinsic Calibration for Event Cameras From First Principles of Events}


\maketitle

\begin{abstract}
The bio-inspired event camera has garnered extensive research attention in recent years, owing to its significant potential derived from its high dynamic range and low latency characteristics.
Similar to the standard camera, the event camera requires precise intrinsic calibration to facilitate further high-level visual applications, such as pose estimation and mapping.
While several calibration methods for event cameras have been proposed, most of them are either ($i$) engineering-driven, heavily relying on conventional image-based calibration pipelines, or ($ii$) inconvenient, requiring complex instrumentation.
To this end, we propose an accurate and convenient intrinsic calibration method for event cameras, named \emph{eKalibr}, which builds upon a carefully designed event-based circle grid pattern recognition algorithm.
To extract target patterns from events, we perform event-based normal flow estimation to identify potential events generated by circle edges, and cluster them spatially.
Subsequently, event clusters associated with the same grid circles are matched and grouped using normal flows, for subsequent time-varying ellipse estimation.
Fitted ellipse centers are time-synchronized, for final grid pattern recognition.
We conducted extensive experiments to evaluate the performance of \emph{eKalibr} in terms of pattern extraction and intrinsic calibration.
The implementation of \emph{eKalibr} is open-sourced at (\url{https://github.com/Unsigned-Long/eKalibr}) to benefit the research community.
\end{abstract}


\section{Introduction}
\IEEEPARstart{T}{he} event camera, as a bio-inspired novel vision sensor,  could overcome the challenges of motion blur and low-illumination degradation encountered by the conventional camera, owing to its low latency and high dynamic range \cite{gallego2020event,huang2023event}.
Leveraging these distinctive characteristics, event cameras have been applied in numerous robotic tasks in recent years, such as pose estimation \cite{huang2023event}, object tracking \cite{mitrokhin2018event}, and structured light 3D scanning \cite{matsuda2015mc3d}, particularly in challenging environments.
To support event-based applications, precise intrinsic calibration is crucial for event cameras, to establish the mathematical mapping between the 3D world and 2D image.
While certain event cameras, such as \emph{DAVIS} series \cite{brandli2014240} and \emph{CeleX} series \cite{chen2019live}, are capable of standard image output, thereby enabling conventional frame-based intrinsic calibration \cite{kalibr}, other event cameras, such as \emph{DVS-Gen} series \cite{suh20201280}, exclusively support event output, leaving their intrinsic calibration an unresolved challenge.


In visual intrinsic calibration, artificial targets, e.g., April Tags \cite{olson2011apriltag} and checkerboards \cite{de2010automatic}, are commonly employed for data association, with accurate target recognition from visual outputs being a critical issue.
For conventional cameras, precise image-based target pattern recognition has been extensively studied and well-established \cite{de2010automatic,wang2016apriltag}.
However, for event cameras, due to their unconventional asynchronous output, image-based pattern recognition methods are not directly applicable, thus extracting accurate target patterns from raw event streams remains a challenging problem that has not been effectively addressed.

\begin{figure}[t]
\centering
\includegraphics[width=\linewidth]{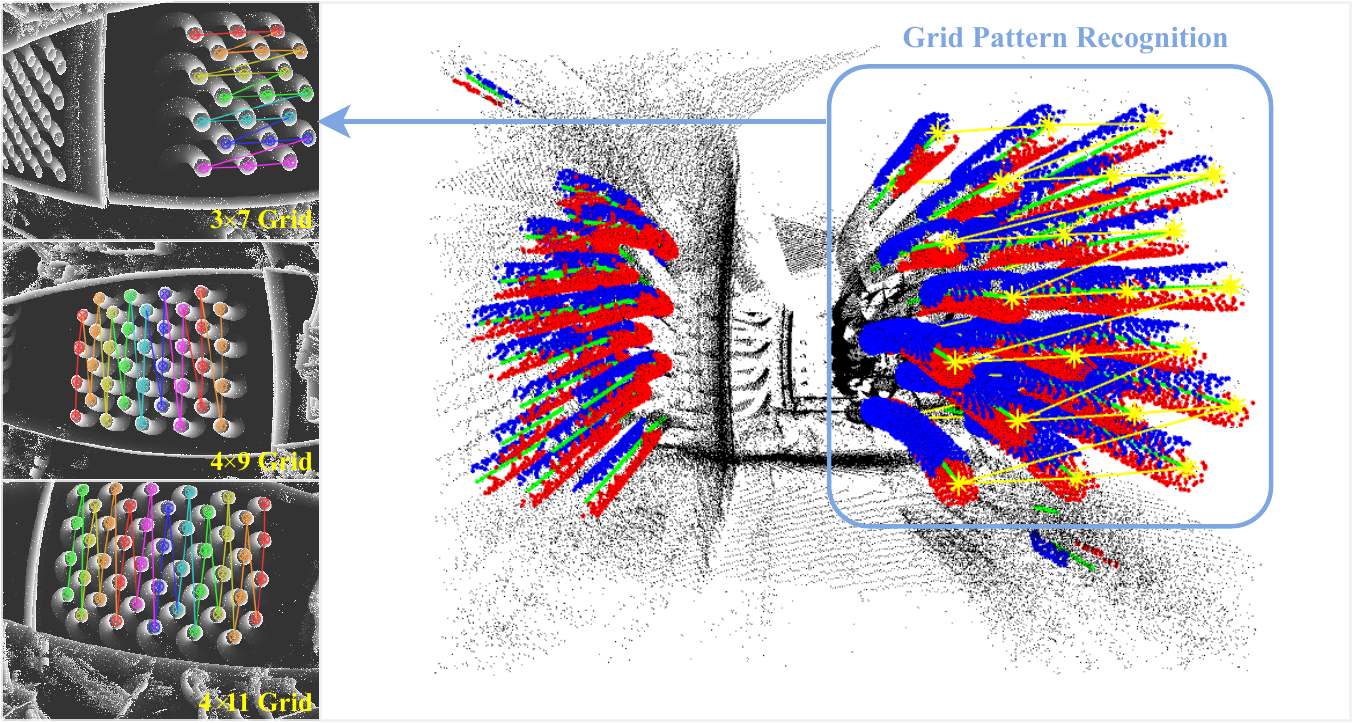}
\caption{The runtime visualization of circle grid pattern recognition in \emph{eKalibr}. \emph{eKalibr} extracts patterns from raw events in the spatiotemporal domain from first principles of events.}
\label{fig:ekalibr}
\figtabbottomspace
\end{figure}

The most straightforward approach for reliably and accurately recognizing target patterns from event streams is to use a blinking light-emitting diode (LED) grid board \cite{gorchard2025dvscalibration,rpg_dvs_ros}.
By accumulating events triggered by the blinking LED board, the grid pattern can be accurately extracted from accumulation images.
Using LEDs for pattern recognition is grounded in first principles of events, easy to understand, and generally could yield accurate calibration results.
However, several inherent limitations accompany this approach:
($i$) it imposes high requirements on the hardware (the LED board),  and ($ii$) a more \textbf{critical} one is that it's a static calibration method, which cannot be used for multi-camera or event-inertial spatiotemporal calibration that requires motion excitation \cite{chen2024ikalibr}.

Another approach for target pattern recognition for event cameras is to directly reconstruct image frames from events and then apply well-established frame-based calibration methods for intrinsic determination \cite{gehrig2021dsec,muglikar2021calibrate}, which is also intuitively understandable and could overcome limitations of LED-based approaches.
However, due to the significant noise present in the reconstructed images, it is challenging to use this method for precise intrinsic calibration.

Considering the issues present in these methods, we propose a rigorous, accurate, and event-based target pattern recognition method oriented to conventional circle grid boards for precise intrinsic calibration of event cameras, named \emph{eKalibr} (see Fig. \ref{fig:ekalibr}).
Specifically, we first perform event-based normal flow estimation to identify inlier events, and then spatially and homopolarly cluster them by contour searching.
Event clusters would be matched based on the prior knowledge from circle-oriented normal flow distribution.
Finally, time-varying ellipses would be estimated using raw events within each event cluster, for synchronous grid pattern extraction from ellipse centers.
\emph{eKalibr} makes the following (potential) contributions:
\begin{enumerate}
\item We propose a dynamic intrinsic calibration method for event cameras from the first principles of events, which leverages the \textbf{common} circle grid board, thus offering both convenience and extensibility.

\item A rigorous and accurate event-based pattern recognition approach oriented to the circle grid board is designed to identify grid patterns from raw events for intrinsic estimation.
This approach has the \textbf{potential} to extend to other event-related calibration, such as event-inertial spatiotemporal calibration.

\item Sufficient experiments were conducted to evaluate the proposed \emph{eKalibr}. Both datasets and code implementations are open-sourced, to benefit the robotic community if possible.
\end{enumerate}

\section{Related Works}

The LED grid board was initially used (also the most widely used in recent years) for event-based target pattern recognition and further intrinsic calibration of event cameras.
Given the fact that the event camera generates events by detecting brightness changes, the authors of \cite{gorchard2025dvscalibration} and \cite{rpg_dvs_ros} employed a blinking LED grid board flashing at a fixed frequency to trigger the event camera, generating low-noise accumulation event images for grid pattern extraction.
Similarly, leveraging LEDs but extracting the target patterns in the phase domain rather than the spatial (intensity) domain, Cai et al. \cite{cai2024accurate} propose a Fourier-transform-based calibration method, which significantly mitigates the impact of event noise during calibration.

Benefiting from the recent advancements in event-to-image algorithms, a new approach for intrinsic calibration of event cameras has emerged, which involves first reconstructing images and then performing image-based intrinsic calibration.
Gehrig et al. \cite{gehrig2021dsec} and Muglikar et al. \cite{muglikar2021calibrate} are the first to apply event-to-image frameworks for event camera calibration. Their approach leverages images constructed by \emph{E2VID} \cite{rebecq2019high} to perform frame-based intrinsic calibration within the \emph{Kalibr} \cite{kalibr}.
Although the impressive performance of event-to-image reconstruction frameworks such as \emph{E2VID} \cite{rebecq2019high} and \emph{EVSNN} \cite{zhu2022event} should be acknowledged, the accuracy of their reconstructed images remains inadequate for precise calibration.

An alternative approach is to directly extract target patterns from raw events for calibration, which is the most rooted in first principles of events, yet is also highly challenging due to the asynchrony, noise, and spatial sparsity of the events \cite{gallego2020event}.
Huang et al. \cite{huang2021dynamic} are the first to focus on recognizing target patterns from events generated by relative motion between the camera and artificial target.
They cluster events generated by a circle grid board using density-based spatial clustering (DBSCAN) for pattern recognition.
Similarly, leveraging DBSCAN, Salah et al. \cite{salah2024calib} propose an efficient reweighted least squares (eRWLS) method to determine event cylinder centers associated with grid circles.
Both \cite{huang2021dynamic} and \cite{salah2024calib} directly cluster accumulated events, making them sensitive to noise and generally resulting in a large number of outlier candidate clusters for grid circles.
Considering this, Wang et al. \cite{wang2024ef} designed a novel circle grid board with cross points for efficient pattern recognition and subsequent joint event-frame spatiotemporal calibration.
A carefully designed event-oriented noise suppression pipeline is also presented in \cite{wang2024ef}.
However, introducing additional cross points in the circle grid board could give rise to potential noise events, thereby negatively affecting circle center determination.

\section{Preliminaries}
This section provides necessary notations and definitions used throughout the article.
The camera intrinsic model and normal flow involved in this work are also introduced, ensuring a self-contained presentation for the reader.
\subsection{Notations and Definitions}
The event camera detects brightness change and generates events at pixels where the intensity difference exceeds a contrast sensitivity \cite{huang2023event}.
We denotes the $j$-th generated event as $\bsm{e}_j$ in this article, which is defined as:
\begin{equation}
\equabovespace
\footnotesize
\bsm{e}_j\triangleq\left\lbrace
\tau_j,\bsm{x}_j,p_j
\right\rbrace
\quad\mathrm{s.t.}\quad
\bsm{x}_j=\begin{bmatrix}
x_j,y_j
\end{bmatrix}^\top\in\mathbb{Z}^2,\;
p_j\in\left\lbrace \smallminus 1,\smallplus 1\right\rbrace
\equbelowspace
\end{equation}
where $\tau_j$ is the time of event $\bsm{e}_j$ stamped by the camera;
$\bsm{x}_j$ denotes the two-dimensional (2D) pixel coordinates where the event locates;
$p_j$ is the polarity of the event, indicating the direction of brightness change.
In terms of coordinate systems, we use $\coordframe{w}$ and $\coordframe{c}$ to represent the world frame (i.e., the coordinate system of the circle grid pattern) and camera frame, respectively.
The three-dimensional (3D) rigid-body transformation from $\coordframe{w}$ to $\coordframe{c}$ is parameterized as the Euclidean matrix:
\begin{equation}
\equabovespace
\footnotesize
\transform{w}{c}\triangleq\begin{bmatrix}
\rotation{w}{c}&\translation{w}{c}\\
\bsm{0}_{1\times 3}&1
\end{bmatrix}\in\mathrm{SE(3)}
\equbelowspace
\end{equation}
where $\rotation{w}{c}\in\mathrm{SO(3)}$ and $\translation{w}{c}\in\mathbb{R}^3$ denote the rotation matrix and translation vector, respectively.
Finally, we represent the noisy measurements and estimated quantities by  $\tilde{(\cdot)}$ and $\hat{(\cdot)}$ respectively.

\subsection{Camera Intrinsic Model}
The camera intrinsic model consists of the projection model and the distortion model, defining the correspondence between 3D objects in the world frame and 2D pixels in the image plane.
Various projection and distortion models exist, such as pinhole \cite{kannala2006generic} and double sphere \cite{usenko2018double} projection models, as well as radial-tangential \cite{tang2017precision} and equidistant (fisheye) \cite{zhou2022orthorectification} distortion models.
In this work, the pinhole projection model and radial-tangential distortion model are considered, and the corresponding camera model can be described as follows:
\begin{equation}
\equabovespace
\label{equ:visual_proj}
\small
\bsm{x}_p=\pi\left( \translation{}{c}, \bsm{x}_{\mathrm{intr}}\right)
\triangleq\begin{bmatrix}
f_x&0&c_x\\
0&f_y&c_y
\end{bmatrix}\cdot
\begin{bmatrix}
x^{\prime\prime}\\y^{\prime\prime}\\1
\end{bmatrix}
\equbelowspace
\end{equation}
with
\begin{equation}
\equabovespace
\footnotesize
\begin{gathered}
\bsm{x}_{\mathrm{intr}}\triangleq\bsm{x}_{\mathrm{proj}}\cup\bsm{x}_{\mathrm{dist}}\\
\bsm{x}_{\mathrm{proj}}\triangleq\left\lbrace
f_x,f_y,c_x,c_y
\right\rbrace,\;
\bsm{x}_{\mathrm{dist}}\triangleq\left\lbrace
k_1,k_2,p_1,p_2
\right\rbrace
\end{gathered}
\equbelowspace
\end{equation}
and
\begin{equation}
\equabovespace
\footnotesize
\begin{aligned}
x^{\prime\prime}&=x^\prime\cdot( 1+k_1\cdot r^2+k_2\cdot r^4)+2p_1\cdot x^\prime\cdot y^\prime+p_2\cdot( r^2+2{x^\prime}^2) \\
y^{\prime\prime}&=y^\prime\cdot( 1+k_1\cdot r^2+k_2\cdot r^4)+2p_2\cdot x^\prime\cdot y^\prime+p_1\cdot( r^2+2{y^\prime}^2) \\
x^\prime&=\bsm{p}^c(x)/\bsm{p}^c(z),\quad
y^\prime=\bsm{p}^c(y)/\bsm{p}^c(z),\quad
r^2={x^\prime}^2+{y^\prime}^2
\end{aligned}
\equbelowspace
\end{equation}
where $\pi(\cdot)$ represents the camera projection function projecting a 3D point $\translation{}{c}$ in $\coordframe{c}$ onto the 2D image plane as $\bsm{x}_p$;
$\bsm{x}_{\mathrm{intr}}$ denotes the camera intrinsics, including projection coefficients $\bsm{x}_{\mathrm{proj}}$ and distortion coefficients $\bsm{x}_{\mathrm{dist}}$;
$f_{x\mid y}$ and $c_{x\mid y}$ denote the focal lengths and principal point respectively, while $k_{1\mid 2}$ and $p_{1\mid 2}$ are radial and tangential distortion coefficients.
The purpose of camera (intrinsic) calibration is to determine $\bsm{x}_{\mathrm{intr}}$.

\begin{figure*}[t]
	\centering
	\includegraphics[width=\figsize\linewidth]{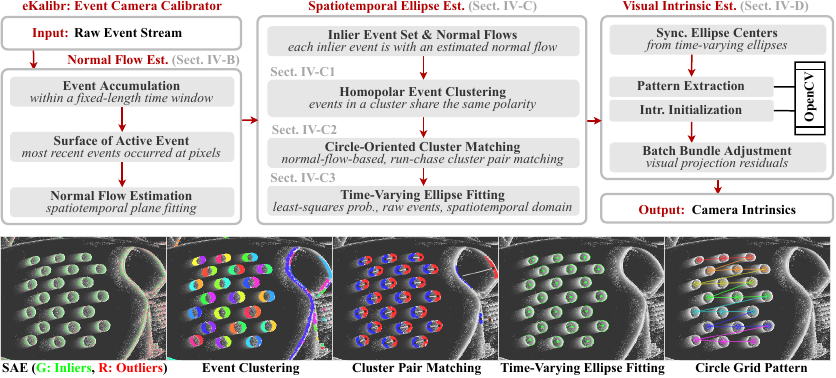}
	\caption{Illustration of the pipeline of the proposed event-based visual intrinsic calibration method. A detailed description of the pipeline is provided in Section \ref{sect:overview}, while detailed methodology is presented in Section \ref{sect:nf_est}, Section \ref{sect:ellipse_est}, and Section \ref{sect:intr_est}.}
	\label{fig:overview}
	\figtabbottomspace
\end{figure*}

\subsection{Normal Flow}


Events are mostly generated by moving high-gradient regions (e.g., edges) in the image \cite{xu2023tight}, and are thus naturally associated with the image gradient.
In the standard vision, the first-order Horn-Schunck model \cite{horn1981determining} gives the following constraint (higher-order items are not considered here):
\begin{equation}
\equabovespace
\footnotesize
\label{equ:horn_schunk}
\mathcal{I}\left( \bsm{x}+\delta\bsm{x},\tau+\delta\tau\right) 
\approx\mathcal{I}\left(\bsm{x},\tau\right) +
\nabla_{\bsm{x}}\mathcal{I}\cdot\delta\bsm{x}+
\nabla_{\tau}\mathcal{I}\cdot\delta\tau
\equbelowspace
\end{equation}
where $\mathcal{I}\left(\bsm{x},\tau\right)$ denotes the image intensity at position $\bsm{x}$ and time $\tau$;
$\nabla_{\bsm{x}}\mathcal{I}$ and $\nabla_{\tau}\mathcal{I}$ are spatial and temporal image gradient, respectively.
Under the assumption of constant image intensity, by dividing (\ref{equ:horn_schunk}) by $\delta\tau$, we obtain:
\begin{equation}
\equabovespace
\footnotesize
\label{equ:optical_flow}
\nabla_{\bsm{x}}\mathcal{I}\cdot\bsm{v}=-
\nabla_{\tau}\mathcal{I}
\quad
\mathrm{s.t.}
\quad
\bsm{v}\triangleq\frac{\mathrm{d}\bsm{x}}{\mathrm{d}\tau}
=\lim_{\delta\tau\to 0}\frac{\delta\bsm{x}}{\delta\tau}
\equbelowspace
\end{equation}
where $\bsm{v}$ is exactly the well-known optical flow (motion flow).
Subsequently, by projecting the optical flow onto the image gradient direction $\nabla_{\bsm{x}}\mathcal{I}$ and introducing (\ref{equ:optical_flow}), we can derive the normal flow $\bsm{n}$ as follows:
\begin{equation}
\equabovespace
\footnotesize
\label{equ:normal_flow}
\bsm{n}\triangleq\frac{
\nabla_{\bsm{x}}\mathcal{I}\cdot\bsm{v}
}{
\Vert\nabla_{\bsm{x}}\mathcal{I}\Vert
}\cdot
\frac{\nabla_{\bsm{x}}\mathcal{I}}{\Vert\nabla_{\bsm{x}}\mathcal{I}\Vert}
=
-\frac{
\nabla_{\tau}\mathcal{I}
}{\Vert\nabla_{\bsm{x}}\mathcal{I}\Vert^2}\cdot\nabla_{\bsm{x}}\mathcal{I}.
\equbelowspace
\end{equation}
The normal flow $\bsm{n}$ can be directly computed from the raw event stream by fitting spatiotemporal planes (see Section \ref{sect:nf_est}), facilitating subsequent circle grid extraction and calibration.

\section{Methodology}
This section presents the detailed pipeline of the proposed event-based visual intrinsic calibration method.

\subsection{Overview}

\label{sect:overview}

The general pipeline of the proposed method is shown in Fig. \ref{fig:overview}.
Given the raw event stream, we first construct the surface of active events (SAE) \cite{delbruck2008frame} within a fixed-length time window, and subsequently estimate normal flows of active events, see Section \ref{sect:nf_est}.
The inlier events in normal flow estimation would be homopolarly clustered (see Section \ref{sect:ev_clustering}), and then one-to-one matched to identify cluster pairs that are generated by the same circle in the grid pattern (see Section \ref{sect:cluster_matching}).
For each cluster pair, we fit a time-varying ellipse to determine the time-continuous curve of the center of the grid circle (see Section \ref{sect:tv_ellipse_fitting}).
Subsequently, time-varying centers would be sampled temporally to the end time of the window to obtain synchronized centers for grid extraction.
Finally, based on extracted circle grid patterns, camera intrinsics can be determined, see Section \ref{sect:intr_est}.


\subsection{Event-Based Normal Flow Estimation}
\label{sect:nf_est}
\begin{figure}[t]
\centering
\includegraphics[width=\figsize\linewidth]{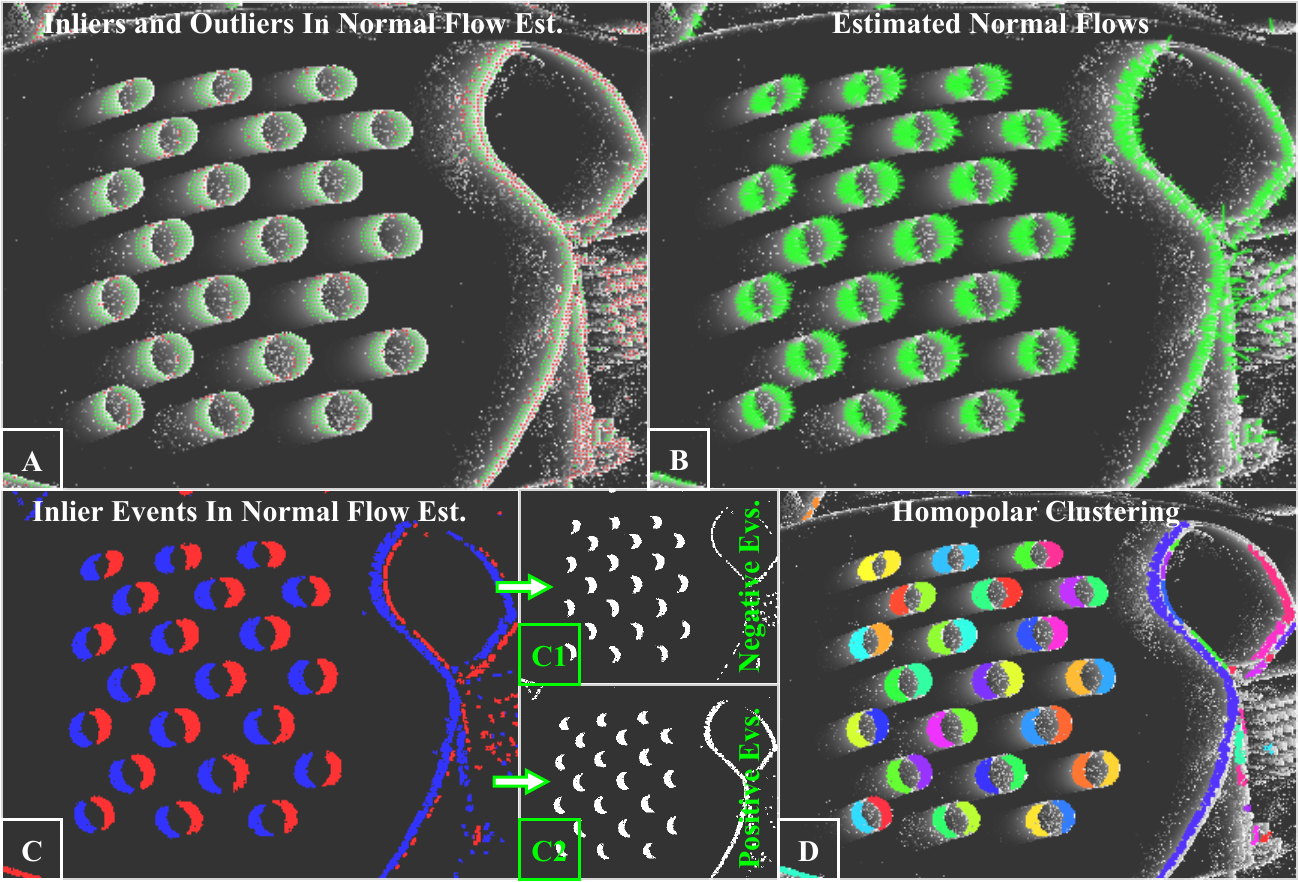}
\caption{Schematic of event clustering.
Subfigure \textbf{A}: events with high (green) and low (red) inlier rates in the plane fitting.
Subfigure \textbf{B}: the estimated normal flows (green lines).
Subfigure \textbf{C}: inlier events, blue ones are positive events (\textbf{C2}) while red ones are negative events (\textbf{C1}).
Subfigure \textbf{D}: clustering results, different colors (randomly generated) represent distinct clusters.
}
\label{fig:nf_cluster}
\figtabbottomspace
\end{figure}

We first perform event-based normal flow estimation for subsequent event clustering and matching.
Given the event stream in a time window of $\Delta\tau$, we accumulate raw events $\mathcal{E}\triangleq\left\lbrace\bsm{e}_j \right\rbrace $ and subsequently construct the time surface map $\mathcal{S}_\mathrm{tm}$ and polarity surface map $\mathcal{S}_\mathrm{pol}$ using $\mathcal{E}$. 
The $\mathcal{S}_\mathrm{tm}$ and $\mathcal{S}_\mathrm{pol}$ record the timestamp and polarity of \textbf{the most recent} event at each pixel respectively, which support direct revisiting of the most recent raw event at a given pixel $\bsm{x}_k$ as follows:
\begin{equation}
\equabovespace
\footnotesize
\begin{gathered}
\mathcal{E}_{\mathrm{srf}}\triangleq\left\lbrace\bsm{e}_k \right\rbrace,
\;\;
\bsm{e}_k\gets\left\lbrace
\mathcal{S}_\mathrm{tm}(\bsm{x}_k),
\bsm{x}_k,
\mathcal{S}_\mathrm{pol}(\bsm{x}_k)
\right\rbrace 
\\
\mathrm{s.t.}\quad
\mathcal{S}_\mathrm{tm}\in\mathbb{R}^{w\times h},\;
\bsm{x}_k\in\mathcal{I},\;
\mathcal{S}_\mathrm{pol}\in\left\lbrace \smallminus 1,\smallplus 1\right\rbrace ^{w\times h}
\end{gathered}
\equbelowspace
\end{equation}
where $w$ and $h$ denote the width and height of the vision sensor;
$\mathcal{E}_{\mathrm{srf}}$ represents active event set lying on the surface.
Subsequently, for each active event, we fit a spatiotemporal plane using its spatial neighboring active events (within a fixed-size window) based on the random sample consensus (RANSAC).
The spatiotemporal plane is parameterized as follows:
\begin{equation}
\equabovespace
\footnotesize
\begin{bmatrix}
x&y&\tau&1
\end{bmatrix}
\cdot\bsm{\Pi}=0
\quad\mathrm{s.t.}\quad
\bsm{\Pi}=\begin{bmatrix}
\Pi_a&\Pi_b&1&\Pi_c
\end{bmatrix}^\top
\equbelowspace
\end{equation}
where $\Pi_a$, $\Pi_b$, and $\Pi_c$ are parameters of plane $\bsm{\Pi}$ to be determined.
The normal flow of this event then can be obtained based on the fitted plane and (\ref{equ:normal_flow}) as follows:
\begin{equation}
\equabovespace
\footnotesize
\bsm{n}=-\frac{1}{\Pi_a^2+\Pi_b^2}\cdot\begin{bmatrix}
\Pi_a\\\Pi_b
\end{bmatrix}
\equbelowspace
\end{equation}
Note that an implicit assumption exists here that the brightness gradient direction is orthogonal to the edges \cite{lu2023event}.
Also, note that only those active events whose associated planes exhibit a high inlier rate in the RANSAC-based plane fitting are selected as inlier events for normal flow computation, see subfigures \{\textbf{A}\} and \{\textbf{B}\} in Fig. \ref{fig:nf_cluster}.
For convenience, we denote the set of inlier events as:
\begin{equation}
\equabovespace
\footnotesize
\mathcal{E}_{\mathrm{inlier}}\triangleq\left\lbrace
\left. \bsm{e}_k\right| \bsm{e}_k\in\mathcal{E}_{\mathrm{srf}},
r_k>r_{\mathrm{thd}}
\right\rbrace 
\equbelowspace
\end{equation}
where $r_k$ denotes the inlier rate of event $\bsm{e}_k$ in the RANSAC-based  spatiotemporal plane fitting. The corresponding normal flow set of $\mathcal{E}_{\mathrm{inlier}}$ is represented as $\mathcal{N}$.

\subsection{Spatiotemporal Ellipse Estimation}
\label{sect:ellipse_est}
Subsequently, event clustering would be performed on obtained $\mathcal{E}_{\mathrm{inlier}}$. The clustered events are then matched as one-to-one pairs for time-varying ellipse fitting.

\subsubsection{Homopolar Event Clustering}
\label{sect:ev_clustering}
Since $\mathcal{E}_{\mathrm{inlier}}$ lies on the surface of active events where no temporal overlap occurs, we conduct clustering in the spatial domain, i.e., the 2D image plane, for efficiency.
Noting that the circle edges generally generate events of two polarities (see Fig. \ref{fig:circle_edge_nfs}), we perform homopolar clustering, i.e., events within a cluster \textbf{share} the same polarity.
To cluster events, the contour searching algorithm \cite{suzuki1985topological} is first employed to identify the contours of event clusters. Events within the same contour are then treated as a single cluster, see subfigure \{\textbf{D}\} in Fig. \ref{fig:nf_cluster}. We denote obtained clusters as:
\begin{equation}
\equabovespace
\footnotesize
\begin{gathered}
\mathcal{C}\triangleq\left\lbrace
\left. \mathcal{C}^k\right| \mathcal{C}^k\gets
\left( \mathcal{E}^k,\mathcal{N}^k\right)
\right\rbrace
\quad\mathrm{s.t.}\quad\\
\mathcal{E}^k\simeq\mathcal{E}^k_{\mathrm{inlier}}\triangleq\left\lbrace
\left. \bsm{e}_j^k\right|\bsm{e}_j^k\in\mathcal{E}_{\mathrm{inlier}}
\right\rbrace,\;
\mathcal{N}^k\triangleq\left\lbrace
\left. \bsm{n}_j^k\right|\bsm{n}_j^k\in\mathcal{N}
\right\rbrace.
\end{gathered}
\equbelowspace
\end{equation}
where $\mathcal{C}^k$ denotes the $k$-th cluster, with event set $\mathcal{E}^k$ and the corresponding normal flow set $\mathcal{N}^k$.

\subsubsection{Run-Chase Cluster Matching}
\label{sect:cluster_matching}
\begin{figure}[t]
	\centering
	\includegraphics[width=\figsize\linewidth]{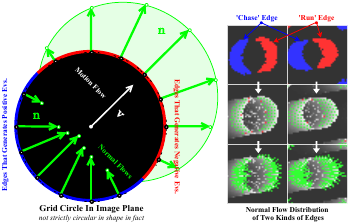}
	\caption{Illustration of the normal flow distribution of edges of grid circles in the image plane. The relative motion between the circle and the camera results in two types of events generated by edges, which exhibit significant distinguishability regarding the directions of the normal flows.}
	\label{fig:circle_edge_nfs}
	\figtabbottomspace
\end{figure}
The relative motion between the circle grid and the camera 
generally results in two event clusters with opposite polarities, see Fig. \ref{fig:circle_edge_nfs}.
Additionally, since the brightness gradient direction is almost orthogonal to the edges of grid circles, the normal flows of events in two clusters associated with a grid circle exhibit distinct distributions.
We intuitively designate the two clusters associated with the same circle as the \textbf{running} cluster and the \textbf{chasing} cluster.
Our objective is to identify running-chasing cluster pairs of potential grid circles from clusters for subsequent time-varying ellipse fitting.

We begin by assigning an initial label of \emph{running}, \emph{chasing}, or \emph{unknown} to each cluster $\mathcal{C}^k$, based on the distribution of normal flow directions.
The following indicator matrix is first computed for each cluster  as the discriminant of the cluster label:
\begin{equation}
\equabovespace
\footnotesize
\bar{\bsm{L}}^k\!=\!\frac{\bsm{L}^k_{\mathrm{avg}}}{\Vert\bsm{L}^k_{\mathrm{avg}}\Vert},\;
\bsm{L}^k_{\mathrm{avg}}=\frac{1}{m}\sum_{j=0}^{m}\begin{bmatrix}
\begin{aligned}
l(d^k_j) \cdot l(s^k_j) & \quad l(d^k_j) \cdot l(-s^k_j) \\
l(-d^k_j) \cdot l(s^k_j) & \quad l(-d^k_j) \cdot l(-s^k_j)
\end{aligned}
\end{bmatrix}
\equbelowspace
\end{equation}
with
\begin{equation}
\equabovespace
\footnotesize
\begin{aligned}
d^k_j &= \bsm{n}^{k}_j \times \bar{\bsm{n}}^{k}, & \;
s^k_j &= \left( \bsm{x}^k_j - \bar{\bsm{x}}^{k} \right) \times \bar{\bsm{n}}^{k},\quad \bar{\bsm{n}}^{k}=\frac{\bsm{n}^{k}_{\mathrm{avg}}}{\Vert\bsm{n}^{k}_{\mathrm{avg}}\Vert},\\
\bsm{n}^{k}_{\mathrm{avg}} &= \frac{1}{m} \sum_{j=0}^{m} \bsm{n}^{k}_j, & \;
\bar{\bsm{x}}^{k} &= \frac{1}{m} \sum_{j=0}^{m} \bsm{x}^{k}_j,\;\;
l(z)\triangleq\begin{cases}
1,&z>0,\\
0,&z\le 0
\end{cases}
\end{aligned}
\equbelowspace
\end{equation}
where $d^k_j$ and $s^k_j$ are values that indicate the location of the normal flow $\bsm{n}^{k}_j$ and position $\bsm{x}^{k}_j$ of an event $\bsm{e}_j^k$, with respect to the average normal flow $\bar{\bsm{n}}^{k}$ and position $\bar{\bsm{x}}^{k}$.
For the ideal cases of running, chasing, and other (unknown) clusters, the indicator matrices have the following forms:
\begin{equation}
\equabovespace
\footnotesize
\bar{\bsm{L}}_{\mathrm{run}}\!=\!\begin{bmatrix}
\sqrt{2}/2&\!\!\!\!\!\!\!\!0\\0&\!\!\!\!\!\!\!\!\sqrt{2}/2
\end{bmatrix},
\bar{\bsm{L}}_{\mathrm{chase}}\!=\!\begin{bmatrix}
0&\!\!\!\!\!\!\!\!\sqrt{2}/2\\\sqrt{2}/2&\!\!\!\!\!\!\!\!0
\end{bmatrix},
\bar{\bsm{L}}_{\mathrm{unk}}\!=\!\begin{bmatrix}
1/2&\!\!\!\!\!1/2\\1/2&\!\!\!\!\!1/2
\end{bmatrix}
\equbelowspace
\end{equation}
Based on this fact, we perform the Frobenius norm-based similarity metric  for each cluster using indicator matrices, and ultimately determine the unique label of each cluster:
\begin{equation}
\equabovespace
\footnotesize
\mathrm{Similarity}\left(\bar{\bsm{L}}^k,\bar{\bsm{L}}_{(\cdot)}\right)\triangleq 1-\frac{
\Vert\bar{\bsm{L}}^k-\bar{\bsm{L}}_{(\cdot)}\Vert_F
}{
\Vert\bar{\bsm{L}}_{(\cdot)}\Vert_F
},
\Vert\bsm{A}\Vert_F\triangleq\sqrt{\sum_{i,j}\vert a_{ij}\vert^2}
\equbelowspace
\end{equation}
where $\Vert\bsm{A}\Vert_F$ denotes the Frobenius norm of matrix $\bsm{A}$.
The label of $\mathcal{C}^k$, denoted as $\mathcal{L}^k$, would be determined as $\mathcal{L}_{\mathrm{run}}$, $\mathcal{L}_{\mathrm{chase}}$, or $\mathcal{L}_{\mathrm{unk}}$ based on the principle of maximum similarity.

\begin{figure}[t]
	\centering
	\includegraphics[width=\figsize\linewidth]{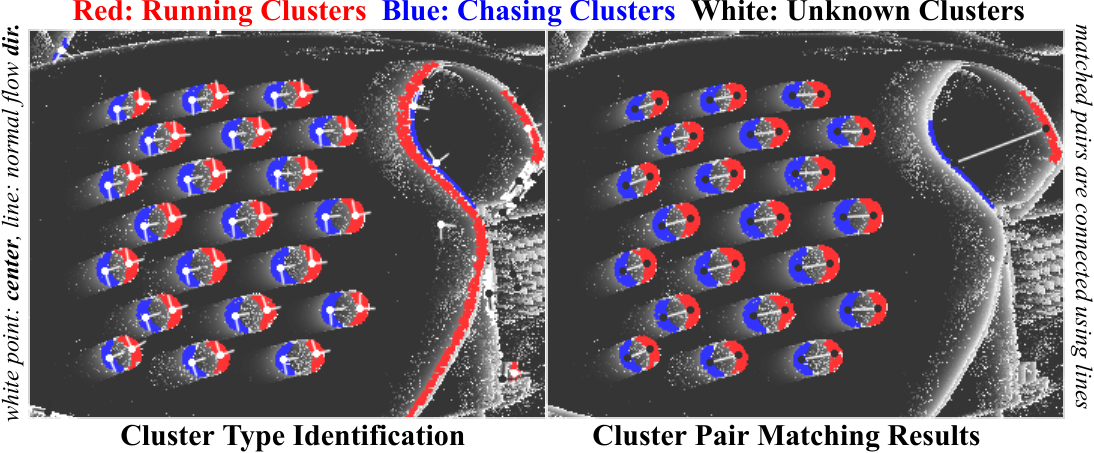}
	\caption{Schematic of cluster type identification (left) and matching (right).}
	\label{fig:cluster_matching}
\end{figure}

Finally, we perform three-stage cluster pair matching, aiming to thoroughly search for potential matching pairs:
($i$) \emph{running-chasing matching}: matching between clusters labeled as $\mathcal{L}_{\mathrm{run}}$ \textbf{and} $\mathcal{L}_{\mathrm{chase}}$;
($ii$) \emph{running/chasing-unknown matching}: matching unmatched $\mathcal{L}_{\mathrm{run}}$ \textbf{or} $\mathcal{L}_{\mathrm{chase}}$ clusters with $\mathcal{L}_\mathrm{unk}$ clusters;
($iii$) \emph{unknown-unknown matching}: matching among unmatched $\mathcal{L}_\mathrm{unk}$ clusters. 
To enhance readers' understanding, the \textbf{first-stage} matching process is summarized in Algorithm \ref{alg:matching}, while cluster matching results are shown in Fig. \ref{fig:cluster_matching}.

\begin{algorithm}[t]
\caption{First-Stage Running-Chasing Cluster Matching}
\label{alg:matching}
\begin{algorithmic}[1]
\State \textbf{Input:} event clusters $\mathcal{C}$ and corresponding (inlier) raw events $\mathcal{E}$, normal flows $\mathcal{N}$, and labels $\mathcal{L}$.
\State \textbf{Output:} One-to-one cluster pairs $\mathcal{P}$.
\For{each cluster $\mathcal{C}^i\in\mathcal{C}$ labeled as $\mathcal{L}_{\mathrm{chase}}$}
\State Initialize candidate cluster $\mathcal{C}^k$, distance $d^{ik}$, index $k$.
\For{each cluster $\mathcal{C}^j\in\mathcal{C}$ labeled as $\mathcal{L}_{\mathrm{run}}$}
\State Compute cluster distance $d^{ij}$ using (\ref{equ:cluster_distance}).
    \If{$d^{ij}<d^{ik}$}
    \State$\mathcal{C}^k\gets\mathcal{C}^j,d^{ik}\gets d^{ij},k\gets j$.
    \EndIf
\EndFor
\State Store the correspondence: $\mathcal{P}\gets\left(\mathcal{C}^i,\mathcal{C}^k,d^{ik}\right)$.
\EndFor
\State Eliminate ambiguous pairs in $\mathcal{P}$ (multiple chasing clusters may be matched to the same running cluster) using the proximity principle, i.e., using stored cluster distance $d^{ik}$.
\State\textbf{Note:} The cluster pair distance $d^{ij}$ is defined as:
\begin{equation}
\equabovespace
\footnotesize
\label{equ:cluster_distance}
d^{ij}\triangleq \mathrm{Distance}\left(\mathcal{C}^i,\mathcal{C}^j\right)=\Vert\bar{\bsm{x}}^{ij}\Vert
\quad\mathrm{s.t.}\quad
\bar{\bsm{x}}^{ij}\triangleq\bar{\bsm{x}}^{j}-\bar{\bsm{x}}^{i}.
\equbelowspace
\end{equation}
If two clusters have the same polarity ($p^i=p^j$), large difference for normal flow directions ($\bar{\bsm{n}}^i\cdot \bar{\bsm{n}}^j <\theta_{\mathrm{thd}}$), or large misalignment ($\bar{\bsm{x}}^{ij}\cdot\bar{\bsm{n}}^{i\mid j}<\theta_{\mathrm{thd}}$), their distance $d^{ij}$ would be set to infinity.
\end{algorithmic}
\end{algorithm}

\subsubsection{Time-Varying Ellipse Fitting}
\label{sect:tv_ellipse_fitting}
After event clustering and matching, raw events within the same cluster pair are grouped for subsequent time-varying ellipse fitting.
Note that our preceding operations (such as clustering and matching) involve only filtering and classification of raw events, thus the \textbf{original} sensor measuring information (raw events) is preserved.

\begin{figure}[t]
	\centering
	\includegraphics[width=\figsize\linewidth]{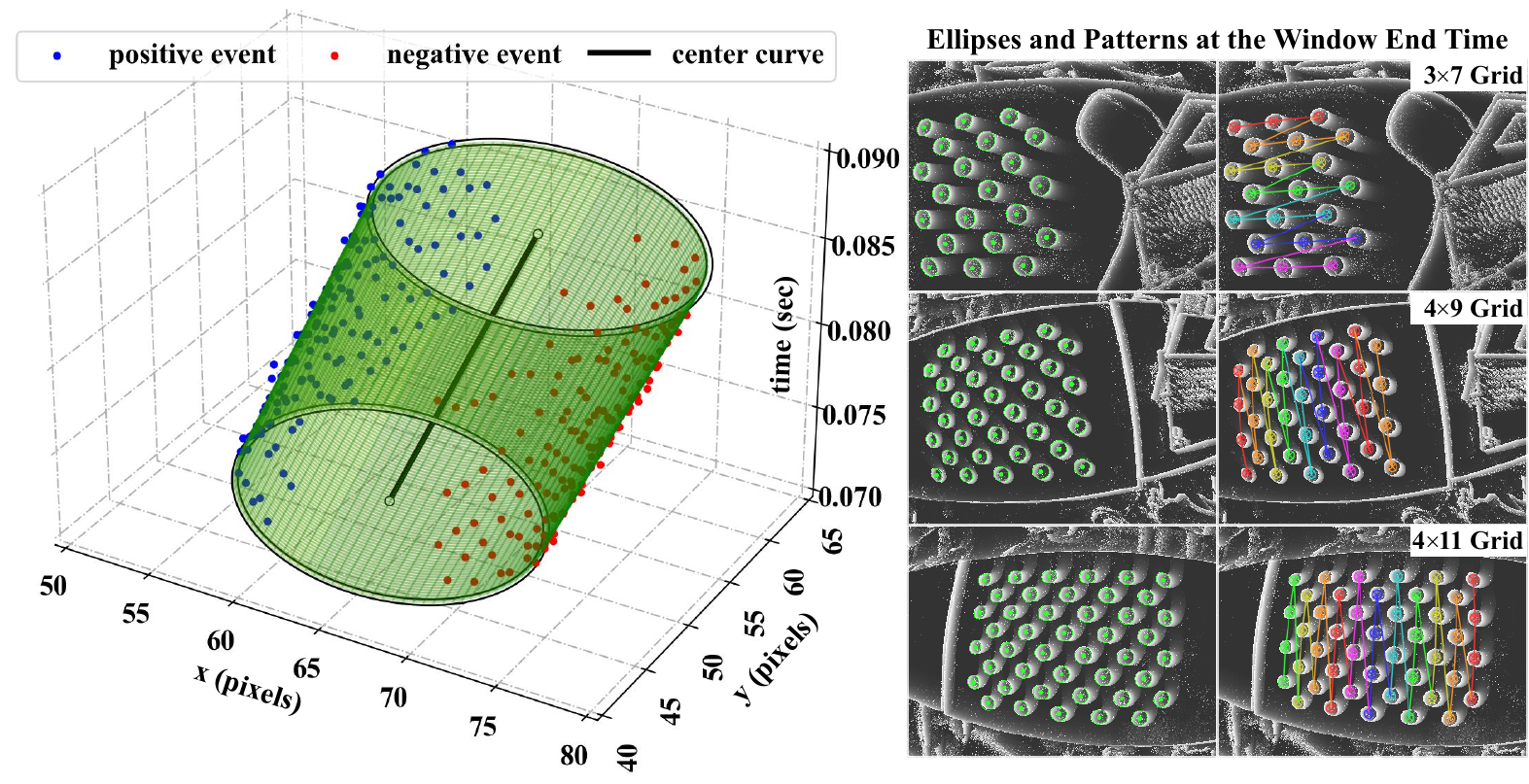}
	\caption{Schematic of time-varying ellipse fitting. The linear time-varying ellipse $E$ (see left subfigure) derived from raw events would be temporally sampled to organize grid patterns (see right subfigures).}
	\label{fig:tv_ellipse}
	\figtabbottomspace
\end{figure}

Due to the oblique perspective and imaging distortions, the projection of a 3D circle onto the 2D image plane often results in a shape that no longer maintains its circular form (also not regular ellipses).
We approximate it using the ellipse for simplicity.
As a result, raw events within a cluster pair are expected to lie on the edges of a time-varying 2D ellipse.
We utilize a linear time-varying ellipse $E$ to model all events within a cluster pair, and parameterize it as:
\begin{equation}
\equabovespace
\footnotesize
E\left(c_{x\mid y}(\tau),\lambda_{x\mid y}(\tau),\alpha\right) :\frac{\left( x^\prime-c_x(\tau)\right)^2}{\left( \lambda_x(\tau)\right) ^2}+
\frac{\left( y^\prime-c_y(\tau)\right)^2}{\left( \lambda_y(\tau)\right) ^2}=1
\equbelowspace
\end{equation}
with
\begin{equation}
\equabovespace
\footnotesize
\bsm{x}^\prime=\bsm{R}\left(\alpha\right) \cdot\bsm{x}
\quad\mathrm{s.t.}\quad
\bsm{x}=\begin{bmatrix}
x&y
\end{bmatrix}^\top,
\bsm{x}^\prime=\begin{bmatrix}
x^\prime&y^\prime
\end{bmatrix}^\top
\equbelowspace
\end{equation}
where $\bsm{R}\left(\alpha\right)\in\mathrm{SO(2)}$ denotes the time-invariant rotation of $E$;
$c_{x\mid y}(\tau)\in P_1$ and $\lambda_{x\mid y}(\tau)\in P_1$ represent the time-varying center and axes of $E$ respectively, all of them are first-degree polynomials in time $\tau$, i.e., $P_1(\tau)=a_0 \cdot\tau+a_1$.
Time-varying ellipse fitting is to determine coefficients of polynomial $c_{x\mid y}(\tau)$ and $\lambda_{x\mid y}(\tau)$, and the rotation angle $\alpha$.
Specifically, for the $i$-th cluster pair $\mathcal{P}^i$, its associated $E^i$ can be estimated by solving the following nonlinear least-squares problem:
\begin{equation}
\equabovespace
\label{equ:tv_ellipse_fit}
\footnotesize
\hat{E}^i\gets
\arg\min\sum_{j=0}^{m}\left\|
r\left( \tilde{\bsm{e}}^i_j\right)
\right\|^2
\equbelowspace
\end{equation}
with
\begin{equation}
\equabovespace
\footnotesize
r\left( \tilde{\bsm{e}}^i_j\right)\triangleq
\hat{\lambda}_y^2(\tau) ( \hat{\tilde{x}}^\prime-\hat{c}_x(\tau))^2+
\hat{\lambda}_x^2(\tau)( \hat{\tilde{y}}^\prime-\hat{c}_y(\tau))^2-\hat{\lambda}_x^2(\tau)\hat{\lambda}_y^2(\tau)
\equbelowspace
\end{equation}
where $r\left( \tilde{\bsm{e}}^i_j\right)$ denotes the residual of event $\bsm{e}^i_j$.
The problem described in (\ref{equ:tv_ellipse_fit}) would be solved in \emph{Ceres} \cite{Agarwal_Ceres_Solver_2022}.
Once $E^i$ is determined, the ellipse at a specific timestamp $\tau$, denoted as $E^i(\tau)$, can be obtained through temporally sampling time-varying $c_{x\mid y}(\tau)$, $\lambda_{x\mid y}(\tau)$, and rotation angle $\alpha$, see Fig. \ref{fig:tv_ellipse}.

\subsection{Visual Intrinsic Estimation}
\label{sect:intr_est}
Finally, we identify circle grid patterns from fitted ellipses, and perform visual batch optimization to obtain final intrinsics.
Specifically, for the $k$-th time window $\left[ \tau_s^k,\tau^k_e\right)$, we perform normal flow estimation, event clustering, cluster pair matching, and time-varying ellipse fitting using in-window raw events.
All fitted time-varying ellipses would be temporally sampled at time point $\tau_e^k$ to obtain synchronous 2D ellipses $\left\lbrace E^i(\tau_e^k)\right\rbrace $.
\textbf{Centers} of 2D ellipses then would be organized as grid pattern using interface \texttt{findCirclesGrid($\cdot$)} in \emph{OpenCV} \cite{opencv_library}, see Fig. \ref{fig:tv_ellipse}.
The found \textbf{ordered} grid pattern in the $k$-th time window is represented as:
\begin{equation}
\equabovespace
\footnotesize
\mathcal{G}^k\triangleq\left\lbrace
\left. \left( \bsm{x}^k_j,\bsm{p}^w_j\right) \right| 
\bsm{x}^k_j=\begin{bmatrix}
c_x^j(\tau_e^k),c_y^j(\tau_e^k)
\end{bmatrix}^\top
\in\mathbb{R}^2,\bsm{p}^w_j\in\mathbb{R}^3
\right\rbrace 
\equbelowspace
\end{equation}
where $\bsm{x}^k_j$ denotes the 2D projection of the 3D center of the $j$-th grid circle (i.e., the $\bsm{p}^w_j$ parameterized in $\coordframe{w}$).

To accurately recover camera intrinsics, we first randomly sample several grid patterns from $\left\lbrace \mathcal{G}^k\right\rbrace$ and compute intrinsic guesses using  \texttt{calibrateCamera($\cdot$)} in \emph{OpenCV}.
The guesses with the lowest root-mean-square error (RMSE) would be selected as initials of intrinsics.
Subsequently, based on intrinsic initials, PnP \cite{lepetit2009ep} is utilized to estimate camera poses $\left\lbrace\transform{w}{c_k}\right\rbrace$.
Finally, a non-linear least-squares batch optimization (bundle adjustment) would be performed to refine all initialized states to global optimal ones, which can be expressed as follows:
\begin{equation}
\equabovespace
\label{equ:batch_opt}
\footnotesize
\hat{\bsm{x}}_{\mathrm{intr}},\left\lbrace\transformhat{w}{c_k}\right\rbrace\gets
\arg\min\sum_{k=0}^{m}\!\!\sum_{j=0}^{g_w\times  g_h}
\rho\left(\left\|
\tilde{\bsm{x}}^k_j-\!\pi\left({\bsm{p}}^{c_k}_j,\hat{\bsm{x}}_{\mathrm{intr}}\right)  
\right\|^2\right)
\equbelowspace
\end{equation}
with
\begin{equation}
\equabovespace
\footnotesize
{\bsm{p}}^{c_k}_j=\rotationhat{w}{c_k}\cdot\bsm{p}^w_j+\translationhat{w}{c_k}
\equbelowspace
\end{equation}
where $\coordframe{c_k}$ is the camera frame associated with $\mathcal{G}^k$; $g_w$ and $g_h$ represent the width (columns) and height (rows) of the grid; $\rho(\cdot)$ denotes the Huber loss function \cite{huber1992robust};
$\pi(\cdot)$ is the visual projection function described in (\ref{equ:visual_proj}).

\section{Real-World Experiment}
This section presents the specific real-world experiments and corresponding results.

\subsection{Experimental Setup}

\begin{figure}[t]
	\centering
	\includegraphics[width=\figsize\linewidth]{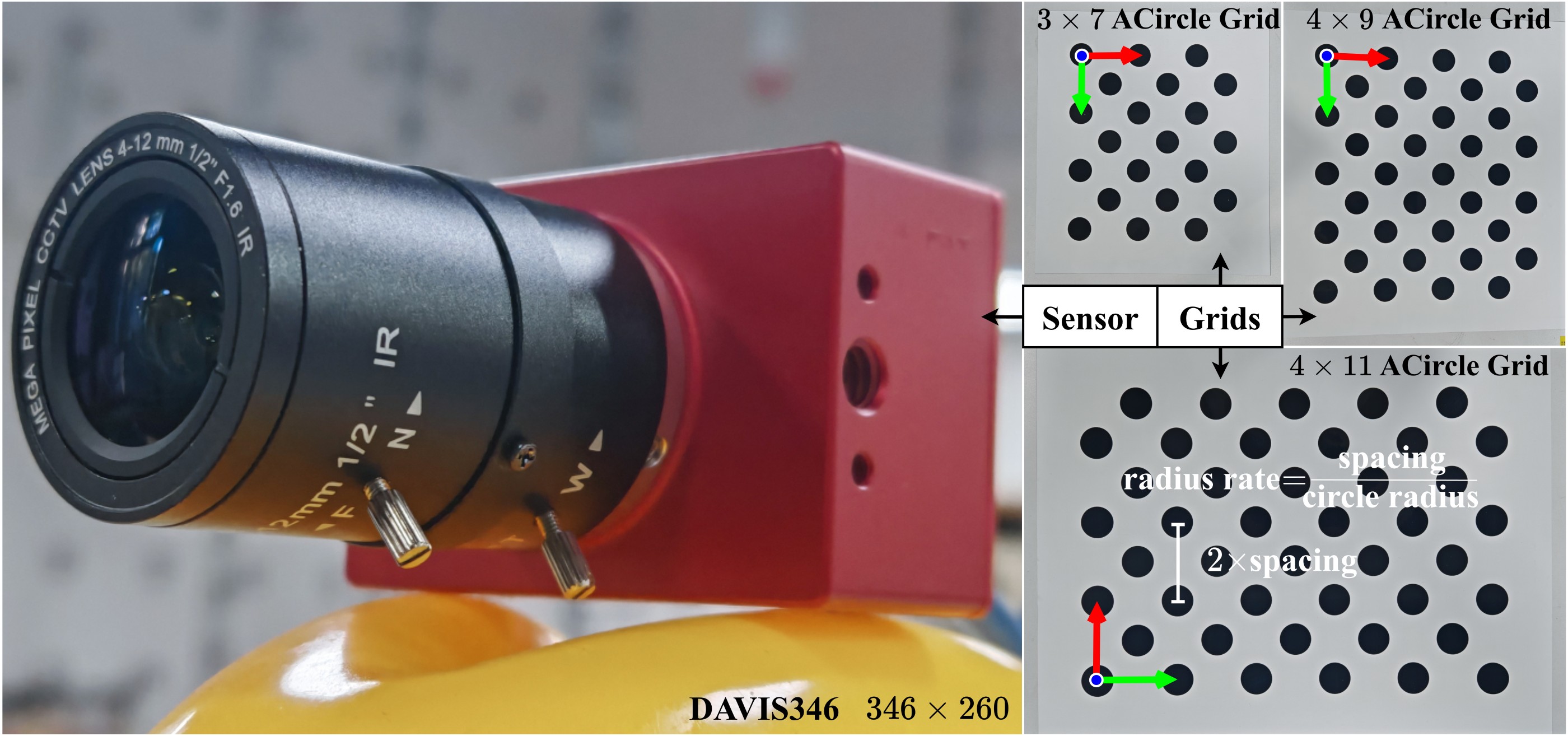}
	\caption{Environmental setup about the equipment (left subfigure) and the circle grid patterns (right subfigures) utilized in real-world experiments.}
	\label{fig:setup}
	\figtabbottomspace
\end{figure}

The event camera, \emph{DAVIS346}, with a resolution of $346\times260$, was employed in our experiments, which supports the acquisition of both raw event streams and conventional image frames.
Asymmetric circle grid patterns\footnote{\textbf{The asymmetric circle pattern} does not exhibit 180-degree ambiguity, making it suitable for more tasks, such as multi-camera extrinsic calibration.} in three different sizes ($3\times7$ for small size, $4\times9$ for medium size, and $4\times11$ for large size) are utilized in our experiments (see Fig. \ref{fig:setup}), to comprehensively evaluate the proposed method.
The spacing and radius rate of three grid patterns are $50$ mm and $2.5$, respectively.
For each pattern, we randomly collected five data sequences for Monte-Carlo experiments, with each having a duration of $30$ sec.
The time window length for the event-based grid pattern recognition is configured to $0.02$ sec.

\subsection{Evaluation and Comparison of Calibration Performance}

\begin{table*}[t]
\renewcommand{\arraystretch}{}
\setlength{\tabcolsep}{\tabwidth}
\centering
\caption{\textbf{Evaluation and Comparison of Intrinsic Calibration Results in Real-World Monte-Carlo Experiments}\\
eKalibr achieves comparable results to the conventional frame-based calibration method
}
\tabtitlespace
\label{tab:calib_res_tab}
\begin{threeparttable}
\begin{tabular}{rr|r|r|r|r}
\hline
\multicolumn{2}{r|}{Parameter}                                                                             & Frame-Based (DV\cite{inivation_dv_tool})             & E2VID \cite{rebecq2019high} + Kalibr \cite{kalibr}   & E-Calib \cite{salah2024calib}                                & eKalibr (Ours)                                                \\ \hline\hline
\multicolumn{1}{r|}{\multirow{7}{*}{\rotatebox{90}{3$\times 7$ Grid}}}  & $f_{x\mid y}$            & 256.47$\pm$1.22, 256.39$\pm$1.25                     & 246.59$\pm$6.78, 247.44$\pm$5.10                     & 255.21$\pm$1.82, 254.09$\pm$1.86                             & 255.30$\pm$\textbf{1.51}, 255.17$\pm$\textbf{1.52}            \\
\multicolumn{1}{r|}{}                                                           & $c_{x\mid y}$            & 169.90$\pm$0.43, 122.17$\pm$0.09                     & 167.36$\pm$1.29, 120.04$\pm$1.12                     & 171.02$\pm$0.58, 121.33$\pm$0.40                             & 170.03$\pm$\textbf{0.29}, 121.96$\pm$\textbf{0.31}            \\ \cline{2-6} 
\multicolumn{1}{r|}{}                                                           & $k_1$                    & -4.31$e^{\smallminus 1}$$\pm$5.47$e^{\smallminus 3}$ & -3.80$e^{\smallminus 1}$$\pm$1.38$e^{\smallminus 2}$ & -4.09$e^{\smallminus 1}$$\pm$5.26$e^{\smallminus 3}$         & -4.27$e^{\smallminus 1}$$\pm$\textbf{4.97$e^{\smallminus 3}$} \\
\multicolumn{1}{r|}{}                                                           & $k_2$                    & 2.84$e^{\smallminus 1}$$\pm$1.86$e^{\smallminus 2}$  & 3.21$e^{\smallminus 1}$$\pm$5.34$e^{\smallminus 2}$  & 2.68$e^{\smallminus 1}$$\pm$\textbf{1.03$e^{\smallminus 2}$} & 2.71$e^{\smallminus 1}$$\pm$1.14$e^{\smallminus 2}$           \\
\multicolumn{1}{r|}{}                                                           & $p_1$                    & 7.56$e^{\smallminus 4}$$\pm$4.37$e^{\smallminus 4}$  & 9.00$e^{\smallminus 4}$$\pm$1.49$e^{\smallminus 3}$  & 3.97$e^{\smallminus 4}$$\pm$6.82$e^{\smallminus 4}$          & 1.85$e^{\smallminus 4}$$\pm$\textbf{5.55$e^{\smallminus 4}$}  \\
\multicolumn{1}{r|}{}                                                           & $p_2$                    & -1.24$e^{\smallminus 2}$$\pm$2.30$e^{\smallminus 2}$ & 8.65$e^{\smallminus 2}$$\pm$2.47$e^{\smallminus 1}$  & 5.06$e^{\smallminus 4}$$\pm$7.01$e^{\smallminus 4}$          & 6.19$e^{\smallminus 4}$$\pm$\textbf{2.86$e^{\smallminus 4}$}  \\ \cline{2-6} 
\multicolumn{1}{r|}{}                                                           & $\sigma_{\mathrm{proj}}$ & 0.07                                                 & 0.35                                                 & 0.21                                                         & \textbf{0.11}                                                 \\ \hline
\multicolumn{1}{r|}{\multirow{7}{*}{\rotatebox{90}{4$\times 9$ Grid}}}  & $f_{x\mid y}$            & 255.41$\pm$1.76, 255.43$\pm$1.77                     & 249.72$\pm$5.05, 248.52$\pm$5.92                     & 255.06$\pm$\textbf{0.98}, 256.89$\pm$1.22                    & 254.57$\pm$1.05, 254.72$\pm$\textbf{1.02}                     \\
\multicolumn{1}{r|}{}                                                           & $c_{x\mid y}$            & 170.21$\pm$0.19, 123.64$\pm$0.34                     & 169.02$\pm$1.04, 121.31$\pm$1.00                     & 173.69$\pm$0.41, 123.61$\pm$0.34                             & 169.44$\pm$\textbf{0.20}, 122.29$\pm$\textbf{0.17}            \\ \cline{2-6} 
\multicolumn{1}{r|}{}                                                           & $k_1$                    & -4.18$e^{\smallminus 1}$$\pm$7.02$e^{\smallminus 3}$ & -3.97$e^{\smallminus 1}$$\pm$9.97$e^{\smallminus 3}$ & -4.29$e^{\smallminus 1}$$\pm$6.22$e^{\smallminus 3}$         & -4.21$e^{\smallminus 1}$$\pm$\textbf{4.40$e^{\smallminus 3}$} \\
\multicolumn{1}{r|}{}                                                           & $k_2$                    & 2.54$e^{\smallminus 1}$$\pm$1.35$e^{\smallminus 2}$  & 3.03$e^{\smallminus 1}$$\pm$4.77$e^{\smallminus 2}$  & 2.68$e^{\smallminus 1}$$\pm$\textbf{6.14$e^{\smallminus 3}$} & 2.52$e^{\smallminus 1}$$\pm$7.79$e^{\smallminus 3}$           \\
\multicolumn{1}{r|}{}                                                           & $p_1$                    & 2.38$e^{\smallminus 4}$$\pm$1.22$e^{\smallminus 4}$  & 7.32$e^{\smallminus 4}$$\pm$1.28$e^{\smallminus 3}$  & 7.93$e^{\smallminus 4}$$\pm$5.30$e^{\smallminus 4}$          & -4.44$e^{\smallminus 4}$$\pm$\textbf{2.14$e^{\smallminus 4}$} \\
\multicolumn{1}{r|}{}                                                           & $p_2$                    & -9.38$e^{\smallminus 2}$$\pm$1.10$e^{\smallminus 2}$ & -1.65$e^{\smallminus 3}$$\pm$1.21$e^{\smallminus 1}$ & 3.63$e^{\smallminus 4}$$\pm$3.39$e^{\smallminus 4}$          & 8.81$e^{\smallminus 4}$$\pm$\textbf{2.82$e^{\smallminus 4}$}  \\ \cline{2-6} 
\multicolumn{1}{r|}{}                                                           & $\sigma_{\mathrm{proj}}$ & 0.06                                                 & 0.37                                                 & 0.28                                                         & \textbf{0.17}                                                 \\ \hline
\multicolumn{1}{r|}{\multirow{7}{*}{\rotatebox{90}{4$\times 11$ Grid}}} & $f_{x\mid y}$            & 255.91$\pm$0.42, 255.87$\pm$0.43                     & 249.08$\pm$4.97, 251.20$\pm$4.70                     & 255.73$\pm$1.03, 255.87$\pm$\textbf{0.91}                    & 255.98$\pm$\textbf{0.98}, 256.10$\pm$1.00                     \\
\multicolumn{1}{r|}{}                                                           & $c_{x\mid y}$            & 170.01$\pm$0.20, 121.73$\pm$0.30                     & 169.63$\pm$0.88, 121.41$\pm$0.56                     & 171.02$\pm$0.63, 122.80$\pm$0.32                             & 169.85$\pm$\textbf{0.23}, 121.73$\pm$\textbf{0.14}            \\ \cline{2-6} 
\multicolumn{1}{r|}{}                                                           & $k_1$                    & -4.23$e^{\smallminus 1}$$\pm$2.02$e^{\smallminus 3}$ & -4.09$e^{\smallminus 1}$$\pm$6.04$e^{\smallminus 3}$ & -4.24$e^{\smallminus 1}$$\pm$4.36$e^{\smallminus 3}$         & -4.23$e^{\smallminus 1}$$\pm$\textbf{3.77$e^{\smallminus 3}$} \\
\multicolumn{1}{r|}{}                                                           & $k_2$                    & 2.70$e^{\smallminus 1}$$\pm$4.87$e^{\smallminus 3}$  & 2.91$e^{\smallminus 1}$$\pm$1.08$e^{\smallminus 2}$  & 2.50$e^{\smallminus 1}$$\pm$7.89$e^{\smallminus 3}$          & 2.54$e^{\smallminus 1}$$\pm$\textbf{6.50$e^{\smallminus 3}$}  \\
\multicolumn{1}{r|}{}                                                           & $p_1$                    & 5.95$e^{\smallminus 4}$$\pm$1.65$e^{\smallminus 4}$  & 7.79$e^{\smallminus 4}$$\pm$2.83$e^{\smallminus 4}$  & 9.05$e^{\smallminus 4}$$\pm$1.83$e^{\smallminus 4}$          & 8.29$e^{\smallminus 4}$$\pm$\textbf{1.62$e^{\smallminus 4}$}  \\
\multicolumn{1}{r|}{}                                                           & $p_2$                    & 6.09$e^{\smallminus 4}$$\pm$4.31$e^{\smallminus 4}$  & 3.84$e^{\smallminus 4}$$\pm$5.67$e^{\smallminus 3}$  & 6.82$e^{\smallminus 4}$$\pm$2.76$e^{\smallminus 4}$          & 6.33$e^{\smallminus 4}$$\pm$\textbf{2.64$e^{\smallminus 4}$}  \\ \cline{2-6} 
\multicolumn{1}{r|}{}                                                           & $\sigma_{\mathrm{proj}}$ & 0.08                                                 & 0.41                                                 & 0.29                                                         & \textbf{0.21}                                                 \\ \hline
\end{tabular}
\begin{tablenotes} 
\item[*] The quantities presented in the row for \( f_{x\mid y} \) (pixels) are arranged in the order of \( f_x \) and \( f_y \). This same ordering applies to \( c_{x\mid y} \) (pixels).
\item[*] The value in each table cell is represented as \textbf{(Mean) \( \pm \) (STD)}. The item with the minimum STD in a row is highlighted in bold. The results in the column \textbf{Frame-Based (DV \cite{inivation_dv_tool})} are considered as ground truth and are excluded from comparison when determining the minimum STD.
\item[*] $\sigma_{\mathrm{proj}}$ denotes the root-mean-square (RMS) reprojection error in intrinsic calibration, unit: pixels.
\end{tablenotes}
\end{threeparttable}
\figtabbottomspace
\end{table*}

The calibration performance of \emph{eKalibr} was first evaluated. To ensure the reliability and comprehensiveness of the evaluation, we selected three state-of-the-art and publicly available calibration methods for comparison with \emph{eKalibr}:
\begin{enumerate}
\item \textbf{Frame-Based (DV\cite{inivation_dv_tool})}:
The calibration toolkit provided by the manufacturer of \emph{DAVIS346} (i.e., \emph{iniVation}). Since it's frame-based, the corresponding calibration results can be considered as the ground truth.

\item \textbf{E2VID \cite{rebecq2019high} + Kalibr \cite{kalibr}}:
The method described in \cite{muglikar2021calibrate}, which utilizes \emph{E2VID} \cite{rebecq2019high} to reconstruct images from raw events, and then employs a frame-based calibration toolkit for intrinsic calibration (we use \emph{Kalibr} \cite{kalibr} in our experiments, as \cite{muglikar2021calibrate} did).

\item \textbf{E-Calib \cite{salah2024calib}}: An event-only intrinsic calibration toolkit using asymmetric circles pattern. The circle grid extraction in \emph{E-Calib} \cite{salah2024calib} is based on density-based
spatial clustering (DBSCAN).
\end{enumerate}

Table \ref{tab:calib_res_tab} summarizes the calibration results of four methods on grid patterns of three different sizes, where the mean and standard deviation (STD) of calibration results from Monte-Carlo experiments are provided.
As can be seen, among the four methods, the conventional frame-based method achieved the best repeatability (smallest STD) with the lowest reprojection error.
This is primarily attributed to well-established image-based pattern recognition algorithms, which can extract accurate patterns from standard images.
Among the other three event-only methods, the \emph{E2VID}-based calibration method exhibits the lowest accuracy, mainly due to the significant noise in reconstructed images, making precise pattern extraction challenging.
As for \emph{eKalibr} and \emph{E-Calib}, \emph{eKalibr} yielded results closest to those frame-based methods in terms of the mean, while also demonstrating better repeatability (as indicated by the bolded values in Table \ref{tab:calib_res_tab}).

\begin{table}[t]
\renewcommand{\arraystretch}{}
\setlength{\tabcolsep}{\tabwidth}
\centering
\caption{\textbf{Evaluation and Comparison of Circle Gird Extraction}\\
eKalibr achieves the highest detection success rate}
\tabtitlespace
\label{tab:grid_extraction}
\begin{threeparttable}
\begin{tabular}{r|ccc}
\hline
\multicolumn{1}{c|}{Method}   & 3$\times 7$ Grid & 4$\times 9$ Grid & 4$\times 11$ Grid \\ \hline\hline
E2VID \cite{rebecq2019high}   & 43.180 \%        & 37.898 \%        & 52.617 \%         \\
E-Calib \cite{salah2024calib} & 59.429 \%        & 68.935 \%        & 70.562 \%         \\
eKalibr (Ours)                & 76.933 \%        & 80.520 \%        & 74.280 \%         \\ \hline
\end{tabular}
\begin{tablenotes} 
\item[*] The detection success rate is obtained by: the number of successful detections divided by the total number of detections.
\end{tablenotes}
\end{threeparttable}
\end{table}

Table \ref{tab:grid_extraction} summarizes the grid detection success rates of three event-based methods on three kinds of grids.
\emph{E-Calib} directly employs DBSCAN clustering method for circle-edge-associated event identification, which exhibits a high dependency on parameters and is sensitive to noise.
In contrast, \emph{eKalibr} utilizes the normal flow estimation to identify inlier events, after which only simple clustering is sufficient to obtain high-quality circle-edge-associated events.
As can be seen in Table \ref{tab:grid_extraction}, \emph{eKalibr} exhibits a higher detection success rate compared to \emph{E2VID} and \emph{E-Calib}.

\subsection{Consistency Evaluation}

\begin{figure}[t]
	\centering
	\includegraphics[width=\linewidth]{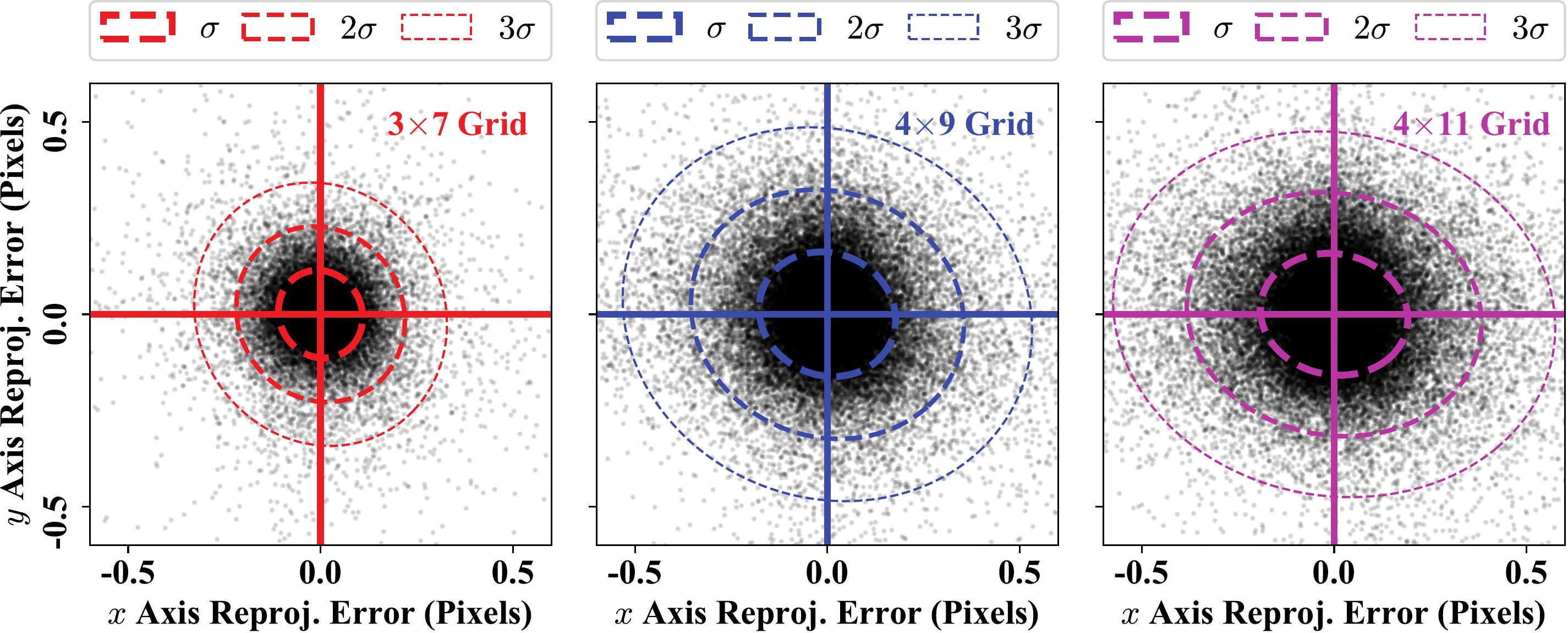}
	\caption{The distribution of reprojection errors in \emph{eKalibr} calibration on three different grids. Solid straight lines represent means of reprojection errors.}
	\label{fig:reproj_error}
	\figtabbottomspace
\end{figure}

\begin{figure}[t]
	\centering
	\includegraphics[width=\figsize\linewidth]{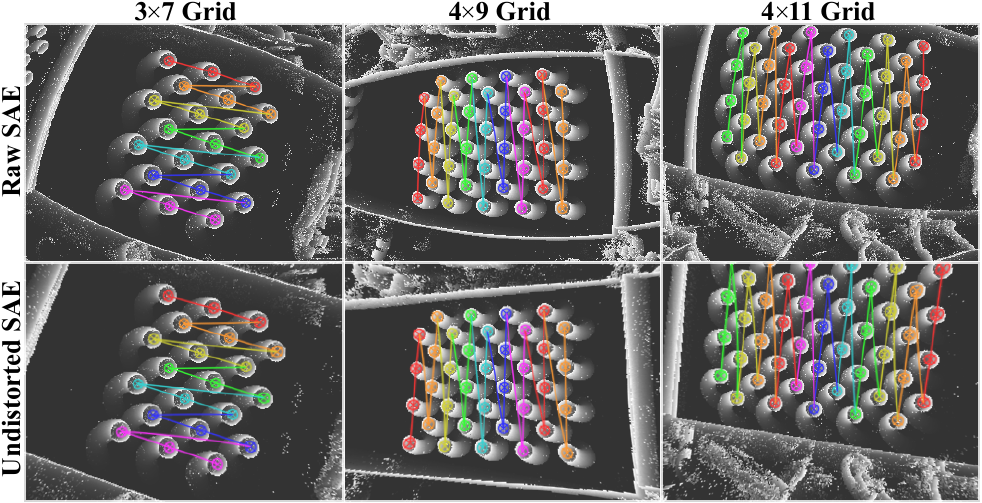}
	\caption{Raw (distorted) SAE maps (top row) and undistorted ones (bottom row) based on intrinsics calibrated by \emph{eKalibr}.}
	\label{fig:undistortion}
\end{figure}

The corresponding distributions of reprojection errors for the RMS values of the proposed \emph{eKalibr} are illustrated in Fig. \ref{fig:reproj_error} for further evaluation of calibration consistency.
The reprojection error represents the difference between the actual image points and the computed ones based on the calibrated intrinsic parameters.
A smaller reprojection error generally indicates a more accurate estimation of the intrinsic parameters.
It can be seen that the reprojection errors of \emph{eKalibr} approximately follow a zero-mean normal distribution, with a small STD of 0.2 pixels (on average), demonstrating its high calibration accuracy.

Fig. \ref{fig:undistortion} illustrates the SAE maps before and after distortion correction using the intrinsics estimated by \emph{eKalibr}.
It can be found that the radial and tangential distortions caused by the lens in the original image have been almost completely eliminated after distortion correction.
The structures in undistorted images align well with their counterparts in the real world, demonstrating a high calibration consistency of \emph{eKalibr}.

\subsection{Computation Consumption Evaluation}

\begin{table}[t]
\renewcommand{\arraystretch}{}
\setlength{\tabcolsep}{\tabwidth}
\centering
\caption{\textbf{Computation Consumption in eKalibr}\\
Grid extraction consumed the majority of the processing time}
\tabtitlespace
\label{tab:computation}
\begin{threeparttable}
\begin{tabular}{c|cclcl}
\hline
\multirow{3}{*}{Config.}               & OS Name                              & \multicolumn{4}{l}{Ubuntu 20.04.6 LTS 64-Bit}                   \\ \cline{2-6} 
                                       & Processor                            & \multicolumn{4}{l}{12th Gen Intel® Core™ i9}                    \\ \cline{2-6} 
                                       & Graphics                             & \multicolumn{4}{l}{Mesa Intel® Graphics}                        \\ \hline\hline
\multirow{2}{*}{Scenes}                & \multicolumn{5}{c}{Computation Consumption (\textbf{unit: minute})}                                    \\ \cline{2-6} 
                                       & \multicolumn{1}{c|}{Grid Extraction} & \multicolumn{2}{c|}{Intrinsic Est.} & \multicolumn{2}{c}{Total} \\ \hline
\multicolumn{1}{r|}{3$\times 7$ Grid}  & \multicolumn{1}{c|}{1.402}           & \multicolumn{2}{c|}{0.047}          & \multicolumn{2}{c}{1.449} \\
\multicolumn{1}{r|}{4$\times 9$ Grid}  & \multicolumn{1}{c|}{2.078}           & \multicolumn{2}{c|}{0.103}          & \multicolumn{2}{c}{2.181} \\
\multicolumn{1}{r|}{4$\times 11$ Grid} & \multicolumn{1}{c|}{2.350}           & \multicolumn{2}{c|}{0.105}          & \multicolumn{2}{c}{2.455} \\ \hline
\end{tabular}
\begin{tablenotes} 
\item[*] The reported time represents the average time consumption across multiple (five) runs, with each data sequence lasting 30 seconds.
\end{tablenotes}
\end{threeparttable}
\figtabbottomspace
\end{table}

Table \ref{tab:computation} summarizes the computational time consumption of \emph{eKalibr} in real-world experiments.
It can be seen that for a 30-second-long dataset, the average time consumption of \emph{eKalibr} is approximately two minutes, with the majority of the time spent on grid extraction.
As the grid size increases, the computational time increases accordingly, due to two primary factors: ($i$) the need to fit more time-varying ellipses using (\ref{equ:tv_ellipse_fit}) for target pattern recognition, and ($ii$) the involvement of more 2D-3D projection correspondences in the final batch optimization described in (\ref{equ:batch_opt}).
Overall, the total computational time is acceptable, demonstrating the high usability of the proposed event-based pattern extraction method. 

\section{Conclusion} 
In this article, we present an open-source visual intrinsic calibration method for event cameras, named \emph{eKalibr}.
Specifically, We first perform event-based normal flow estimation to filter out potential events generated by circle edges.
The filtered events are then clustered in the spatial domain to obtain event clusters for circle-oriented one-to-one matching.
Each matched cluster pair is regarded as corresponding to the same potential grid circle, and would be utilized for time-varying ellipse estimation.
Finally, temporally synchronized grid patterns would be extracted from ellipse centers for final visual intrinsic calibration.
We conduct sufficient experiments to evaluate the proposed \emph{eKalibr}, and the results demonstrate that \emph{eKalibr} is capable of accurate grid pattern extraction and intrinsic calibration.
Compared to existing methods based on LED or event-to-image approaches, \emph{eKalibr} ($i$) employs common visual targets, offering both convenience and extensibility, ($ii$) can extract accurate patterns from dynamically collected raw events for intrinsic determination, and ($iii$) offers the advantages of efficiency, high accuracy, and high repeatability.
In future work, we will support multi-camera and event-inertial spatiotemporal calibration in \emph{eKalibr}.


\section*{CRediT Authorship Contribution Statement}
\label{sect:author_contribution}
\textbf{Shuolong Chen}: Conceptualisation, Methodology, Software, Validation, Original Draft.
\textbf{Xingxing Li}: Supervision, Funding Acquisition.
\textbf{Liu Yuan} and \textbf{Ziao Liu}: Data Curation, Review and Editing.
	
\bibliographystyle{IEEEtran}
\bibliography{reference}

\end{document}